\documentclass[letterpaper, 10 pt, conference]{ieeeconf}  

\makeatletter
\let\NAT@parse\undefined
\makeatother

\usepackage[svgnames]{xcolor}

\usepackage{graphicx}
\usepackage{marvosym}
\usepackage{lipsum}
\usepackage{booktabs}
\usepackage{multirow}
\usepackage{amsmath}
\usepackage{amssymb}
\usepackage{cite}
\usepackage{makecell}
\usepackage{subfiles}

\usepackage[colorlinks=true, linkcolor=LightSlateGrey, citecolor=LightSlateGrey, urlcolor=LightSlateGrey]{hyperref}

\IEEEoverridecommandlockouts                              
\begin{document}
\title{\LARGE \bf
DF$^3$: World Modeling via Decoder-Free Feature Forecasting in Autonomous Navigation
}

\newcommand{\versionmode}{1}

\ifcase\versionmode
  \author{\textbf{Anonymous Submission}}
\or
  \author{Jiaming Chen$^{1}$, Guoan Xu$^{2}$, Aoshen Huang$^{3}$, Haozhuo Zhang$^{1}$, Yang Li$^{\text{\Letter}4}$, Wei Pan$^{\text{\Letter}1}$
  \thanks{$^{\text{\Letter}}$ Corresponding authors.}
  \thanks{$^{1}$Jiaming Chen, Haozhuo Zhang, and Wei Pan are with the Department of Computer Science, The University of Manchester, Manchester M13
9PL, United Kingdom.
        {(email: \tt\footnotesize wei.pan@manchester.ac.uk})}%
\thanks{$^{2} $Guoan Xu is with the Faculty of Engineering and Information Technology, University of Technology Sydney,
Sydney, NSW 2007, Australia.}
\thanks{$^{3} $Aoshen Huang is with the School of Control Science and Engineering,
Shandong University, Jinan 250061, China.}
\thanks{$^{4} $Yang Li is with the School of Computer Science, Shanghai Jiao Tong
University, Shanghai 200240, China. (email: \tt\footnotesize yang.li.cs@sjtu.edu.cn)}
}
\fi

\maketitle
\thispagestyle{empty}
\pagestyle{empty}


\begin{abstract}
  Forecasting future states from video sequences is a critical challenge for autonomous robotic systems and a fundamental objective of world modeling. Prior generative methods operating at the pixel level inevitably overemphasize task-irrelevant details, leading to prohibitive computational overhead. While latent-based approaches attempt to mitigate this by predicting features directly, the persistent reliance on heavy decoders for state-to-task mapping remains a computational bottleneck. In this work, we propose Decoder-Free Feature Forecasting (DF$^3$), a novel framework that models world evolution entirely within the latent space and directly derives task outputs, completely eliminating the need for a decoder. Specifically, DF$^3$ injects learnable spatial queries into the terminal blocks of a frozen vision foundation model to extract future state representations directly. By employing a lightweight, unified Motion-Aware Context Fusion (MACF) mechanism that seamlessly integrates coarse flow warping with fine-grained latent cross-correlation, these queries interact with historical token representations to explicitly align and forecast the feature of the next frame. Subsequently, a specialized set of task queries probes these forecasted features for the downstream task. Extensive experiments on public benchmarks and zero-shot deployment in a robotic simulator demonstrate that DF$^3$ achieves performance comparable to state-of-the-art methods while offering superior efficiency and flexibility for integrated perception and control. 
\end{abstract}

\section{Introduction}

Forecasting future states from video sequences serves as a cornerstone of world modeling and enables autonomous systems to plan safely in complex environments~\cite{zhu2024sora,zhao2025pwm,xiong2026unidrive}. Driven by the emergence of powerful video generators, a family of generative methods performs future-state prediction directly in the raw pixel space~\cite{bar2025navigation,agarwal2025cosmos,ren2025cosmos}. However, video generation imposes a heavy computational burden. Furthermore, utilizing these pixel-level predictions for downstream applications inevitably necessitates an additional perception stage where a task decoder is required to post-process and translate the synthesized video into task outputs~\cite{russell2025gaia,li2026papnet,ando2023rangevit}. More critically, for decision-making tasks such as autonomous navigation, the objective is to capture scene dynamics at a semantic level to support planning and control, rather than to synthesize photorealistic future frames~\cite{zheng2025world4drive,wang2026drive}. 

Latent-based methods~\cite{dinowm,karypidis2025dinoforesight} offer an alternative paradigm by predicting states in the semantic feature space instead of the pixel domain. Leveraging vision foundation models (VFMs)~\cite{clip,tschannen2025siglip,dinov2} that provide semantically rich representations, they forecast features capturing scene dynamics, which are then decoded into task outputs such as semantic segmentation for decision-making. This approach reduces focus on irrelevant details and lowers computational cost. However, predicting high-dimensional latent features poses a new bottleneck. Since the feature space is more abstract and harder to supervise, prior latent-based methods~\cite{dinowm,karypidis2025dinoforesight} often require heavy decoders for contextual modeling. In addition, an extra task-specific decoder is still needed to translate predicted features into task outputs, introducing further computational overhead and limiting their applicability to real-world scenarios.

Given the strong semantic representation capabilities of modern VFMs, this work investigates a fundamental question: \emph{\textbf{Can feature forecasting be performed entirely within the encoder space of a VFM?}} \textcolor{black}{Recent advances have demonstrated the potential of adapting pretrained vision encoders for future state prediction. A representative example is EoMT~\cite{eomt}, which repurposes plain Vision Transformer (ViT)~\cite{vit} architectures to perform dense prediction tasks like semantic segmentation without the need for dedicated decoders.} To this end, we propose the Decoder-Free Feature Forecasting (\textbf{DF$^3$}) framework, which forecasts future state purely within a pre-trained ViT. The core idea is to employ learnable queries that are injected into the terminal blocks of the ViT, enabling interaction with historical context to produce intermediate features for future frames. Leveraging the powerful ViT that benefits from pre-training paradigm such as DINO~\cite{dinov2,dinov3}, these intermediate features remain semantically rich and can support another set of task queries, which are also injected into the ViT to extract task-specific representations for downstream applications. This design eliminates the need for heavy decoders, resulting in a lightweight and efficient architecture well-suited for real-world tasks where compute and latency are critical.

However, this decoder-free design presents a key challenge: a frozen vision encoder, originally developed for static image encoding, lacks the mechanisms to capture temporal dependencies across historical frames and align moving objects in the latent space~\cite{videomt}. To address this issue, we propose a lightweight \textcolor{black}{Motion-Aware Context Fusion (MACF) mechanism}. Rather than relying on a massive spatio-temporal transformer decoder, our approach introduces a unified motion-aware fusion module that seamlessly integrates coarse flow warping with fine-grained latent cross-correlation to explicitly model motion between historical frames. This explicitly aligns the temporal context, allowing our injected queries to accurately probe and forecast future token representations without fine-tuning the pre-trained weights of the encoder.

The proposed DF$^3$ framework is illustrated in Fig.~\ref{fig:pipeline}. DF$^3$ is designed to be highly flexible, as prediction queries and task queries can be trained independently and utilized efficiently. The prediction queries focus purely on forecasting future states in the latent space, while the task queries are tailored to specific downstream objectives like semantic segmentation. We extensively validate our framework on Cityscapes~\cite{cityscapes} benchmark and demonstrate its practical zero-shot deployment for autonomous navigation within simulation. Experimental results show that DF$^3$ maintains competitive forecasting accuracy compared to state-of-the-art latent models while significantly reducing inference latency and peak GPU memory, thereby unlocking real-time capabilities for autonomous robotic applications.

The core contributions can be summarized as follows: 
\begin{itemize} 
\item We propose DF$^3$, a novel decoder-free feature forecasting framework that bypasses the computational bottlenecks of pixel-level and latent decoding. To the best of our knowledge, this is the first work to perform future forecasting entirely within the frozen feature space of a vision foundation model.
\item We introduce a query-based interaction mechanism equipped with a unified, lightweight motion-aware context fusion mechanism that integrates flow warping and latent cross-correlation. This design enables a frozen vision encoder to effectively capture complex temporal dynamics without finetuning its pre-trained weights.
\item We validate DF$^3$ on a forecasting benchmark and demonstrate its zero-shot deployment within a quadruped robot simulator. Results show that our method achieves competitive forecasting accuracy while significantly improving efficiency, offering a highly efficient and practical solution for real-time robotic perception.
\end{itemize}

\section{Related Work}
\label{sec:related_work}
In this section, we first review recent advances in \emph{vision foundation models} that empower large-scale visual representation learning. We then summarize progress in the development of \emph{world models} for predictive understanding and future forecasting. Finally, we discuss the evolution of \emph{efficient and query-based vision architectures}, which provides the structural inspiration for our decoder-free framework design.
 
\subsection{Vision Foundation Models}

Vision foundation models have emerged as large-scale pre-trained networks that learn general-purpose visual representations applicable across diverse tasks~\cite{awais2025foundation,firoozi2025foundation}. Early works like CLIP~\cite{clip} leveraged weakly supervised signals to learn transferable visual embeddings. More recently, self-supervised learning at scale has demonstrated remarkable potential. Particularly, DINOv2~\cite{dinov2} demonstrated that with a sufficiently large and curated dataset and model, self-supervised pretraining can yield features rivaling or surpassing supervised ones. Building on this, DINOv3~\cite{dinov3} has pushed visual foundation models to unprecedented scale and performance by training ViT on an enormous dataset ($\approx$1.7 billion images) using a 7 billion-parameter teacher model. While existing research primarily utilizes these foundation models as feature extractors for static images, our work explores their potential for modeling temporal dynamics and future forecasting.

\subsection{World Models and Future Forecasting}

World modeling aims to construct an internal representation of environment dynamics to anticipate future states, which is essential for planning and control.

\noindent\textbf{Pixel-space Forecasting.} Driven by advances in deep generative models, recent pixel-level forecasting methods~\cite{sora,cogvideox,gao2024vista,harvey2022flexible,zheng2024genad} heavily rely on predicting future states directly in the raw pixel space. While these approaches synthesize visually plausible and high-fidelity video sequences, operating in the pixel domain imposes an extreme computational burden. Furthermore, pixel-level generation inevitably forces the model to allocate significant capacity to task-irrelevant, high-frequency details (e.g., texture, lighting)~\cite{fu2024exploring,dang2025sparseworld}. More critically, deploying these generated videos for downstream robotic tasks necessitates an additional perception stage to decode the synthesized pixels back into actionable semantic outputs, introducing further latency and complexity.

\noindent\textbf{Latent-space Forecasting.} To mitigate the inefficiencies of pixel-level generation, an alternative paradigm performs forecasting within abstract, latent feature spaces~\cite{vjepa,dreamerv3,xiao2025learning,lienhancing}. Recent works such as DINO-WM~\cite{dinowm} and DINO-Foresight~\cite{karypidis2025dinoforesight} capitalize on the semantic richness of vision foundation models, predicting future features rather than raw images. Although this approach effectively filters out irrelevant pixel-level noise, predicting high-dimensional latent representations presents its own computational challenges. Existing latent world models typically rely on heavy Transformer-based spatio-temporal decoders to align context and reconstruct future features~\cite{dinoworld}. Additionally, translating these forecasted features into downstream decisions still requires an external task-specific decoder, maintaining a persistent computational bottleneck for real-time applications.

\subsection{Efficient and Query-based Vision Architectures}

The shift towards efficient visual parsing has been significantly accelerated by query-based architectures. Pioneered by DETR~\cite{detr} and advanced by models like Mask2Former~\cite{mask2former}, these approaches utilize a set of learnable queries to interact with image features, directly extracting object or semantic representations without the need for complex, hand-crafted post-processing.
More recently, researchers have explored pushing this efficiency further by entirely eliminating the decoder stage. Notably, the Encoder-only Mask Transformer (EoMT)~\cite{eomt} demonstrated that a plain, frozen Vision Transformer can be re-purposed for dense prediction tasks simply by injecting task queries into its terminal blocks, effectively bypassing the need for a dedicated decoder. 

\begin{figure}[t]
\centering
\includegraphics[width=0.8\linewidth]{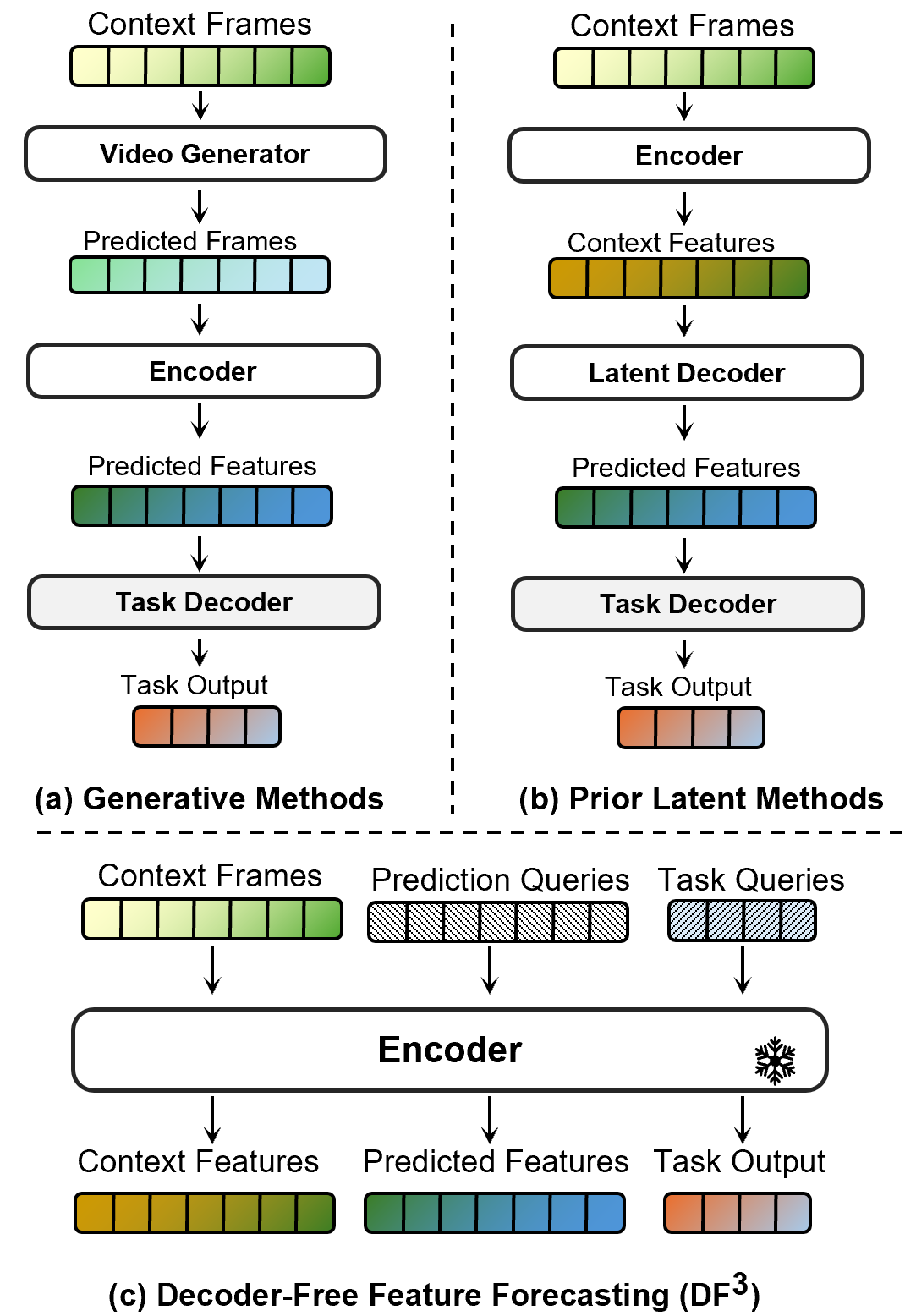}
\caption{Conceptual comparison of forecasting paradigms. (a) \textbf{Generative methods} perform heavy computation in pixel space, generating redundant visual details. (b) \textbf{Prior latent methods} predict semantic features but rely on massive feature and task decoders. (c) Our \textbf{DF$^3$ framework} operates entirely within a frozen vision encoder. By employing query injection and a lightweight motion-aware context fusion module, it forecasts future states and derives task outputs in a fully decoder-free manner. }
\label{fig:teaser}
\end{figure}

\noindent\textbf{Distinctions from Prior Work. }
Inspired by the decoder-free philosophy of EoMT~\cite{eomt}, our proposed DF$^3$ framework represents the first attempt to introduce query-based, decoder-free mechanisms into complex feature forecasting. To clarify our contributions, as conceptually compared in Fig.~\ref{fig:teaser}, we distinguish DF$^3$ from existing methods in two key aspects. First, unlike pixel-level generative methods that suffer from generating redundant visual details, our approach predicts future states directly in a semantically rich feature space. Second, in contrast to prior latent methods that rely on heavy decoders, DF$^3$ forecasts future states and derives downstream task outputs entirely within the frozen encoder space, thereby eliminating the computational bottleneck of decoders. 

\begin{figure*}[ht]
    \centering
    \includegraphics[width=0.9\linewidth]{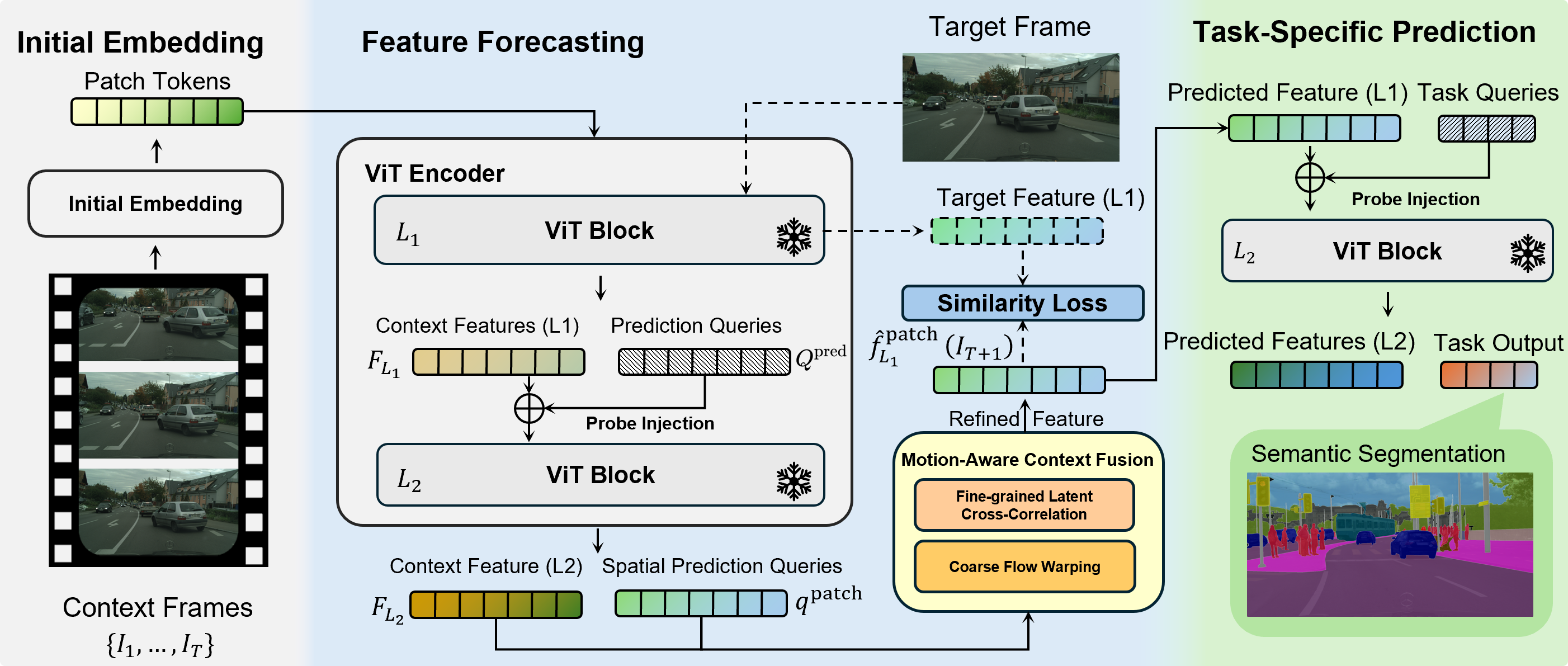}
    \caption{The overall pipeline of the proposed DF$^3$ framework.}
    \label{fig:pipeline}
\end{figure*}

\section{Method}
\label{sec:method}

The overall framework of DF$^3$ is illustrated in Fig.~\ref{fig:pipeline}. Given a sequence of historical context frames, our goal is to forecast future semantic features of the next frame directly within the latent space of the frozen ViT encoder. These features serve as a world representation that can be efficiently shared across downstream tasks. In this section, we first introduce the overall DF$^3$ pipeline in Sec.~\ref{subsec:pipeline}, elaborate on the structured query attention in Sec.~\ref{subsec:attn_mask}, detail our motion-aware context fusion mechanism in Sec.~\ref{subsec:context}, and finally present the training objectives in Sec.~\ref{subsec:objectives}.

\subsection{DF$^3$ Pipeline}
\label{subsec:pipeline}

Let $\{I_{1}, \ldots, I_{T}\}$ denote a sequence of context frames from historical observation, where each $I_t \in \mathbb{R}^{H \times W \times 3}$. We employ a pre-trained DINO-based ViT encoder $E$, which remains entirely frozen during training. As shown in Fig.~\ref{fig:pipeline}, each frame $I_t$ is first passed through the \emph{initial embedding}, including patch embedding and positional encoding, producing a sequence of patch tokens that serve as the input to the subsequent ViT blocks. Since the final layer of $E$ yields features that are highly abstract and lack the dense spatial structure needed for pixel-level prediction, we therefore use the output of earlier layers of $E$, denoted as $L_1$, as our context representation. For each frame $I_t$, $L_1$ yields both patch and prefix tokens including the global class token and register tokens~\cite{dinov2}, which are denoted as the context features:
\begin{equation}
    F_{L_1}(I_t) = \{{f^{\text{pref}}_{L_1}}(I_t), {f^{\text{patch}}_{L_1}}(I_t)\},
\end{equation}
where ${f^{\text{pref}}_{L_1}} \in \mathbb{R}^{N_{\text{pref}} \times C}$ represents the prefix tokens, and ${f^{\text{patch}}_{L_1}} \in \mathbb{R}^{H'W' \times C}$ denotes the patch embeddings with spatial resolution $H'\!\times\!W'$ in feature space.

The core objective is to predict the feature representation $\hat{F}_{{L_1}}(I_{T+1})$ of the next frame $I_{T+1}$. To achieve this, we introduce a set of learnable prediction queries
$Q^{\text{pred}} \in \mathbb{R}^{(H'W' + N_\text{pref}) \times C}$.
Instead of relying on a separate, computationally heavy spatio-temporal decoder, these queries are injected directly into the \emph{terminal blocks} of the frozen encoder, denoted as $L_2$. The prediction queries, concatenated with the context features $\{{F}_{{L_1}}(I_{1}), \ldots, {F}_{{L_1}}(I_{T})\}$, are then processed by $L_2$. The outputs corresponding to the prediction queries are further refined by our motion-aware context fusion mechanism (Sec.~\ref{subsec:context}), and then reshaped into prefix tokens $\hat{f}^{\text{pref}}_{L_1}(I_{T+1}) \in \mathbb{R}^{N_{\text{pref}} \times C}$ and patch-level features $\hat{f}^{\text{patch}}_{L_1}(I_{T+1}) \in \mathbb{R}^{H'W' \times C}$, which together form the final forecasted feature $\hat{F}_{L_1}(I_{T+1})$ of the next frame. 



To derive downstream task-specific predictions, following the EoMT~\cite{eomt}, we perform downstream task prediction by probe-injecting a set of learnable \emph{task queries} $Q^{\text{task}}$ into the \emph{terminal blocks} ($L_2$) of the same frozen ViT encoder. Concretely, the forecasted feature $\hat{F}_{L_1}(I_{T+1})$ is treated as the representation of the next frame, and the task queries are concatenated with $\hat{F}_{L_1}(I_{T+1})$ and fed through the same terminal blocks. Through self-attention, the task queries attend to the forecasted tokens and aggregate task-relevant information. The output tokens corresponding to $Q^{\text{task}}$ are then passed through lightweight MLP heads to produce task-specific outputs. For semantic segmentation, these heads yield mask and class logits that are combined into a dense segmentation map. Crucially, this design keeps the entire pipeline decoder-free since both forecasting and task prediction are performed inside the encoder via query injection. 

\subsection{Structured Query Attention}
\label{subsec:attn_mask}

To enforce a clear separation between context modeling and future prediction, we apply an asymmetric attention mask inside the terminal ViT blocks where the queries are injected. Let $N_q$ denote the total number of prediction queries and $N_c$ denote the number of historical context tokens. The sequence input to the self-attention layer is constructed as $X = [Q^{\text{pred}}, F_{L_1}(I_1), \dots, F_{L_1}(I_T)] \in \mathbb{R}^{(N_q + N_c) \times C}$. We define a boolean attention mask $M \in \{0, 1\}^{(N_q + N_c) \times (N_q + N_c)}$ such that for any query index $i$ and key index $j$:
\begin{equation}
M_{i,j} = 
\begin{cases} 
0, & \text{if } i \ge N_q \text{ and } j < N_q \\
1, & \text{otherwise}
\end{cases}
\end{equation}
where $M_{i,j} = 0$ masks out the attention logit to $-\infty$. This mask permits context tokens ($i \ge N_q$) to attend freely to one another ($j \ge N_q$), but strictly prevents them from attending back to the prediction queries ($j < N_q$). Conversely, the prediction queries ($i < N_q$) can attend to all tokens ($j \ge 0$). This causal design prevents information leakage and ensures that the context representations remain uncorrupted by the injected queries during the spatial-temporal interaction.

\subsection{Motion-Aware Context Fusion Mechanism}
\label{subsec:context}

A central challenge in DF$^3$ is equipping a frozen image encoder, originally designed for static image inputs, with the ability to capture temporal dynamics. While the structured query attention mechanism aggregates global temporal context, explicitly capturing local motion trajectories in the abstract feature space requires dedicated modeling. To achieve this without altering the pre-trained weights, we design a unified Motion-Aware Context Fusion (MACF) mechanism that seamlessly integrates coarse flow warping with fine-grained latent cross-correlation. For a target future frame at time $T+1$, let $q^{\text{patch}} \in \mathbb{R}^{H'W' \times C}$ denote the intermediate spatial prediction queries output from the terminal ViT blocks, and $f^{\text{patch}}_{L_1}(I_T), f^{\text{patch}}_{L_1}(I_{T-1})$ denote the frozen spatial features of the most recent historical context frames. Our module dynamically refines the prediction queries through two parallel branches:

\noindent\textbf{1) Flow-based Warp Branch:} We first predict a dense latent flow field $\Delta p \in \mathbb{R}^{H' \times W' \times 2}$ from the spatial prediction queries:
\begin{equation}
    \Delta p = \text{MLP}_{\text{flow}}(q^{\text{patch}}).
\end{equation}
This continuous flow field models the macroscopic spatial displacement to the next frame. We then warp the most recent historical feature map $f^{\text{patch}}_{L_1}(I_T)$ using this flow, yielding a coarsely aligned future feature $\tilde{f}_{\text{warp}}$. Rather than replacing the query features entirely, we extract a proposed update direction by measuring the difference between the warped feature and the current query state:
\begin{equation}
    M_{\text{warp}} = \text{MLP}_{\text{warp}}(\tilde{f}_{\text{warp}} - q^{\text{patch}}).
\end{equation}

\noindent\textbf{2) Historical Motion via Latent Cross-Correlation:} While the warp branch proposes updates based on the queried future, the cross-correlation branch explicitly extracts true historical velocity to guide the prediction. We compute a local soft-matching to align the previous feature $f^{\text{patch}}_{L_1}(I_{T-1})$ to the latest feature $f^{\text{patch}}_{L_1}(I_T)$. For each spatial location $i$ in $f^{\text{patch}}_{L_1}(I_T)$, we compute the cosine similarity with features in a local $(2r+1)\times(2r+1)$ neighborhood window $\mathcal{N}_r(i)$ in $f^{\text{patch}}_{L_1}(I_{T-1})$. The resulting historical semantic motion $V_{\text{hist}}$ is defined as the difference between the actual feature $f^{\text{patch}}_{L_1}(I_T)$ and the soft-matched previous feature:
\begin{equation}
\begin{split}
    V_{\text{hist}} &= f^{\text{patch}}_{L_1}(I_T) \\
    &\quad - \sum_{k \in \mathcal{N}_r(i)} \frac{\exp(\text{sim}_k / \tau)}{\sum_{j \in \mathcal{N}_r(i)} \exp(\text{sim}_j / \tau)} f^{\text{patch}}_{L_1}(I_{T-1})^{(k)},
\end{split}
\end{equation}
where $\tau$ is a learnable temperature; $\text{sim}_k$ is the cosine similarity between the feature at location $i$ in $f^{\text{patch}}_{L_1}(I_T)$ and the feature at neighbor $k$ in $f^{\text{patch}}_{L_1}(I_{T-1})$. This historical velocity is then projected to form a semantic motion injection:
\begin{equation}
    M_{\text{xcorr}} = \text{MLP}_{\text{xcorr}}(V_{\text{hist}}).
\end{equation}

\noindent\textbf{3) Dynamic Fusion:} Finally, the proposed warp update and the historical motion injection are concatenated and fused. A learned sigmoid gating mechanism $G$ dynamically modulates how much the initial spatial query should trust this fused residual:
\begin{equation}
    \hat{f}^{\text{patch}}_{L_1}(I_{T+1}) = q^{\text{patch}} + G(q^{\text{patch}}) \odot \text{MLP}_{\text{fuse}}\left([M_{\text{warp}}, M_{\text{xcorr}}]\right).
\end{equation}

Additionally, for global prefix tokens, we apply a lightweight temporal extrapolator. It predicts the evolution of global semantics based on the historical difference of the intermediate prefix queries at consecutive time steps, injecting this extrapolated motion to yield the final $\hat{f}^{\text{pref}}_{L_1}(I_{T+1})$. This comprehensive fusion strategy explicitly aligns the temporal context and provides strong inductive biases for modeling complex dynamics.

\subsection{Training Objectives}
\label{subsec:objectives}

Unlike traditional latent world models that rely on joint training with downstream decoders to shape their latent representations, DF$^3$ is trained entirely via feature-level supervision. This isolates the world modeling process, forcing the prediction queries to learn generalizable, task-agnostic dynamics. Once trained, these queries can be seamlessly coupled with task-specific queries to support various downstream applications.

The primary objective aligns the predicted future feature $\hat{F}_{L_1}(I_{t+1})$ with the ground-truth frozen encoder feature $F_{L_1}(I_{t+1})$ for all context timesteps $t \in \{1, \dots, T\}$. To ensure robust learning in the high-dimensional latent space, we combine a cosine similarity term to encourage directional alignment and a Huber loss~\cite{huber} term to stabilize the magnitude:
\begin{equation}
\begin{split}
    \mathcal{L}_{\text{sim}} &= \mathbb{E}_t \left[ 1 - \cos\bigl(\hat{F}_{L_1}(I_{t+1}), F_{L_1}(I_{t+1})\bigr) \right. \\
    &\quad \left. + \lambda_{\text{huber}} \text{Huber}\bigl(\hat{F}_{L_1}(I_{t+1}), F_{L_1}(I_{t+1})\bigr) \right].
\end{split}
\end{equation}
We use $\mathcal{L}_{\text{sim}}$ as the training objective and compute it for both the patch tokens and the prefix tokens to ensure holistic scene representation.

\section{Experiments}
\label{sec:experiments}


In this section, we evaluate the proposed DF$^3$ framework on the Cityscapes dataset~\cite{cityscapes} and demonstrate its deployment within MATRiX~\cite{zsibot2026matrix}, an autonomous navigation simulator for quadruped robots. We first outline the experimental setup and implementation details. Subsequently, we present quantitative comparisons against baseline methods, efficiency comparison against prior latent forecasting methods, and extensive ablation studies to validate the effectiveness of our core components. Qualitative visualizations of the predicted features and segmentation results are provided for both Cityscapes and MATRiX. 

\begin{table*}[h]
    \centering
    \caption{\textbf{Performance and efficiency comparison on the Cityscapes validation set.} We evaluate Short-term ($T+1$) and Mid-term ($T+3$) forecasting performance with mIoU and MO-mIoU. Efficiency comparison includes FLOPs (GFLOPs), latency (ms), peak GPU memory (GB), and forecast parameters (MB). 
    }
    \label{tab:main_results}
    \footnotesize
    \setlength{\tabcolsep}{4.5pt}
    \begin{tabular}{l|cc|cc|llll }
        \toprule
        & \multicolumn{2}{c|}{\textbf{Short} ($t{+}1$)} & \multicolumn{2}{c|}{\textbf{Mid} ($t{+}3$)} & \multicolumn{4}{c}{\textbf{Efficiency}} \\
        Method & mIoU & MO-mIoU & mIoU & MO-mIoU & FLOPs$^\downarrow$ & Latency (ms)$^\downarrow$ & Memory (GB)$^\downarrow$ & Params (MB)$^\downarrow$ \\
        \midrule
        \textcolor{gray}{\textit{Oracle (Upper Bound)}} & \textcolor{gray}{79.8} & \textcolor{gray}{79.7} & \textcolor{gray}{79.8} & \textcolor{gray}{79.7} & \multicolumn{4}{c}{\textcolor{gray}{--}} \\
        \addlinespace
        \midrule
        DINO-Foresight~\cite{karypidis2025dinoforesight} & \textbf{71.8} & \textbf{71.7} & \textbf{59.8} & \textbf{57.6} & 2256.07 & 971.1 & 9.5 & 302.50 \\
        \midrule
        \textbf{DF$^3$ (Ours)} & 69.9 & 68.7 & 58.2 & 56.5 & \textbf{1440.63} \textcolor{gray}{\scriptsize (-36\%)} & \textbf{292.4} \textcolor{gray}{\scriptsize (-70\%)} & \textbf{3.1} \textcolor{gray}{\scriptsize (-67\%)} & \textbf{177.0} \textcolor{gray}{\scriptsize (-41\%)} \\
        \bottomrule
    \end{tabular}
\end{table*}

\begin{table}[h]
    \centering
    \caption{\textbf{Ablation of Context Fusion Strategies.} Evaluating the effectiveness of different temporal aggregation modules.}
    \label{tab:ablation_fusion}
    \footnotesize
    \begin{tabular}{l|cc}
        \toprule
        Fusion Mode & mIoU & MO-mIoU \\
        \midrule
        Concat & 59.9 & 58.0 \\
        Attn & 63.0 & 57.8 \\
        Warp & 63.6 & 60.3 \\
        Xcorr & 65.7 & 63.9 \\
        \textbf{Warp-Xcorr (DF$^3$)} & \textbf{69.9} & \textbf{68.7} \\
        \bottomrule
    \end{tabular}
\end{table}

\begin{table}[h]
    \centering
    \caption{\textbf{Impact of Xcorr Search Radius.} Larger radii help capture fast motion.}
    \label{tab:radius}
    \footnotesize
    \begin{tabular}{l|cc}
        \toprule
        Radius Set & mIoU & MO-mIoU \\
        \midrule
        $r=\{1\}$ & 66.3 & 63.7 \\
        $r=\{1, 2\}$ & 66.8 & 62.2 \\
        $r=\{1, 2, 4\}$ & 69.4 & 66.6 \\
        $\mathbf{r=\{1, 2, 4, 8\}}$ & \textbf{69.9} & \textbf{68.7} \\
        $r=\{1, 2, 4, 8, 16\}$ & 69.7 & 68.6 \\
        \bottomrule
    \end{tabular}
\end{table}

\subsection{Experimental Setup}

\noindent\textbf{Dataset.} We train the DF$^3$ model and conduct comparison experiments on the Cityscapes~\cite{cityscapes} dataset, a large-scale benchmark for semantic urban scene understanding. It contains 2,975 training, 500 validation, and 1,525 test video sequences. Each sequence consists of 30 frames, with the $20^{th}$ frame annotated with fine-grained semantic labels. Following evaluation protocols as DINO-Foresight~\cite{karypidis2025dinoforesight}, we evaluate on two settings: (1) \textit{Short-term Forecasting}, where we use a sequence length of $T=5$ frames to predict the features of the immediate next frame ($T+1$); and (2) \textit{Mid-term Forecasting}, where we autoregressively predict features for multiple future steps (up to $T+3$) to assess temporal stability.

\noindent\textbf{Implementation Details.}
Our framework is built upon the pre-trained DINOv3-ViT-B/16~\cite{dinov3} backbone. The backbone weights are kept frozen to strictly evaluate the forecasting capability within the encoder space. We train the model for 400 epochs using the AdamW~\cite{adamw} optimizer with a base learning rate of $5 \times 10^{-5}$ and a cosine annealing schedule with linear warmup. The input image resolution is $768 \times 1536$, processed via a sliding window of $768 \times 768$.
For the context fusion, we set the cross-correlation search radius to $r \in \{1, 2, 4, 8\}$ to capture multi-scale motion.
The loss weights are set to  $\lambda_{\text{huber}}=0.5$, $\lambda_{\text{nce}}=0.5$. We train the DF$^3$ on a cluster with 4 NVIDIA A100 GPUs, test and deploy it on a server with a single NVIDIA RTX 5090 GPU.

\noindent\textbf{Evaluation Metrics.}
We evaluate the quality of the forecasted features using two complementary sets of metrics:
\begin{enumerate}
    \item \textit{Downstream Performance:} As described in Sec.~\ref{subsec:pipeline}, we use EoMT-style task queries to obtain semantic segmentation from the forecasted features. We report \textit{mIoU} and \textit{MO-mIoU} (Moving Object mIoU on dynamic classes such as car, rider, person).
    \item \textit{Efficiency:} We report \textit{FLOPs} (GFLOPs per frame), \textit{Inference Latency} (ms per frame), \textit{Peak GPU Memory} (GB), and \textit{Parameters} (M) to compare computational cost against prior latent forecasting methods.
\end{enumerate}

\subsection{Comparison Results and Efficiency}
\label{subsec:comparison}

We compare DF$^3$ against the prior latent forecasting method, DINO-Foresight~\cite{karypidis2025dinoforesight}, and an Oracle upper bound that utilizes ground-truth features. As shown in Table~\ref{tab:main_results}, DF$^3$ delivers competitive forecasting accuracy, achieving 69.9 mIoU and 68.7 MO-mIoU for short-term predictions, while maintaining robust mid-term performance. Although DINO-Foresight yields slightly higher accuracy, DF$^3$ effectively offsets this minor drop with massive architectural efficiency. By leveraging a decoder-free design, our method tackles the primary bottleneck of real-time deployment. 

Specifically, on the computational side, DF$^3$ requires $36\%$ fewer GFLOPs and significantly reduces latency by \textbf{$70\%$} (cutting processing time down to just 292 ms per frame). On the spatial side, our method is highly lightweight as it decreases peak GPU memory usage by \textbf{$67\%$} and reduces the total forecast parameters by $41\%$. This substantial reduction in both computational and memory overhead demonstrates that DF$^3$ provides a highly favorable accuracy-efficiency trade-off, making it particularly well-suited for autonomous systems with strict hardware constraints.

\subsection{Ablation Studies}

\noindent\textbf{Context Fusion Strategies.}
We ablate the temporal aggregation module that refines the prediction queries using historical context. In each variant, we keep the same query injection and attention mask and only replace the fusion mechanism after the terminal blocks. We compare the following schemes:
\begin{itemize}
    \item \textit{Concat}: Concatenates the queries with stacked context patch features and passes them through a fusion MLP, with no explicit motion modeling.
    \item \textit{Attn}: Lets the queries attend to all context tokens via self-attention.
    \item \textit{Warp}: Uses only the flow branch.
    \item \textit{Xcorr}: Uses only the cross-correlation branch. 
    \item \textbf{Warp-Xcorr} (DF$^3$): Uses both branches with learned gating.
\end{itemize}

Table~\ref{tab:ablation_fusion} reports short-term ($T+1$) mIoU and MO-mIoU of this ablation.
Flat reaches 59.9 mIoU and 58.0 MO-mIoU, Attn 63.0 and 57.8, Warp 63.6 and 60.3, and Xcorr 65.7 and 63.9. The full Warp-Xcorr mechanism achieves 69.9 mIoU and 68.7 MO-mIoU. Naive concatenation and global attention alone are insufficient for this task. Adding explicit motion through either warp or xcorr brings clear gains, and combining both is the best. Warping handles coarse displacement and occlusion, while cross-correlation refines local alignment and fast motion.

\noindent\textbf{Search Radius in Cross-Correlation.}
The search radius $r$ in the cross-correlation module defines the local window size used for soft-matching consecutive frames. While a larger radius allows the model to capture faster motions, it also increases the susceptibility to spatial noise and false matches. We ablate different radius configurations and report the resulting short-term metrics in Table \ref{tab:radius}.

With restricted radius sets such as $r=\{1\}$ and $r=\{1, 2\}$, the model yields suboptimal performance because a small matching window cannot resolve motions that span multiple patches. Gradually adding larger scales steadily improves alignment. For instance, expanding the set to $r=\{1, 2, 4\}$ notably increases the mIoU to 69.4. Extending the configuration further to $r=\{1, 2, 4, 8\}$ achieves the best overall results of 69.9 mIoU and 68.7 MO-mIoU. The specific inclusion of radius 8 allows the correlation branch to capture fast-moving objects like vehicles at high relative speeds. This capability is particularly important for maximizing the MO-mIoU metric.
However, using an even larger set with $r=\{1, 2, 4, 8, 16\}$ causes a slight performance drop. We hypothesize that the extra radius of 16 enlarges the matching neighborhood beyond what is useful for a single-step prediction. This excessive range likely dilutes the correlation signal and increases sensitivity to background noise. Therefore, we adopt $r=\{1, 2, 4, 8\}$ in all subsequent experiments.

\begin{figure*}
    \centering
    \includegraphics[width=0.9\linewidth]{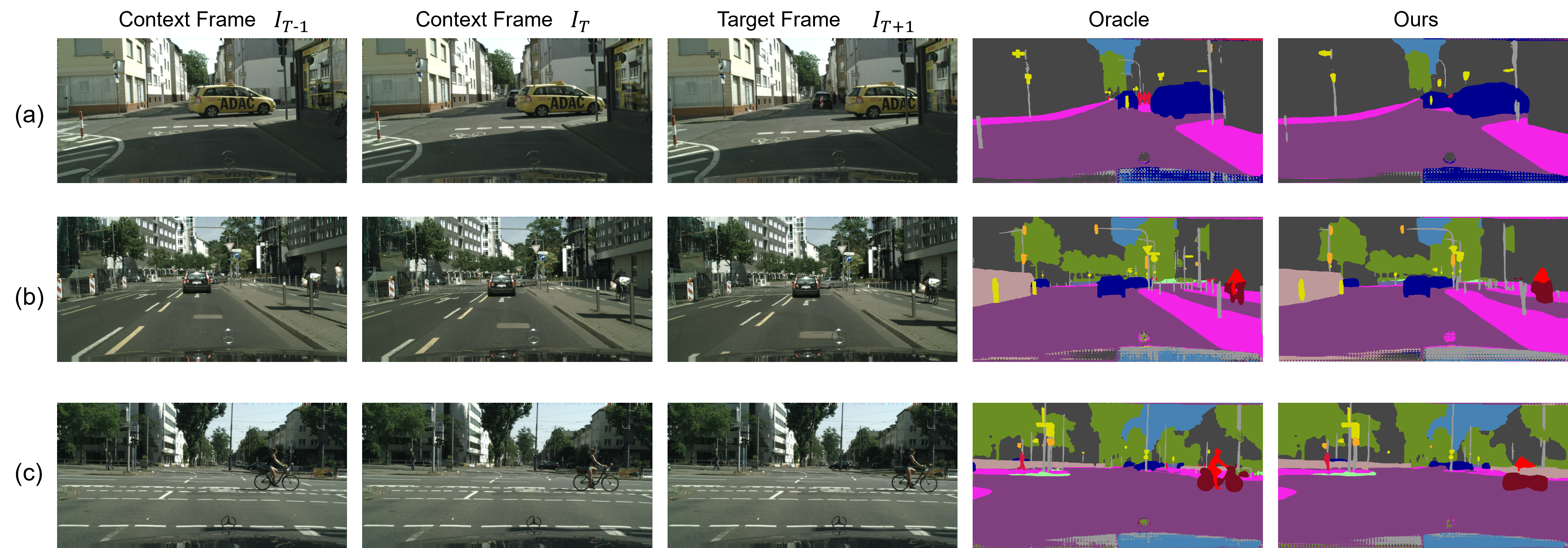}
    \caption{The visualization results of DF$^3$ on Cityscapes. }
    \label{fig:vis_seg}
\end{figure*}

\begin{figure*}
    \centering
    \includegraphics[width=0.9\linewidth]{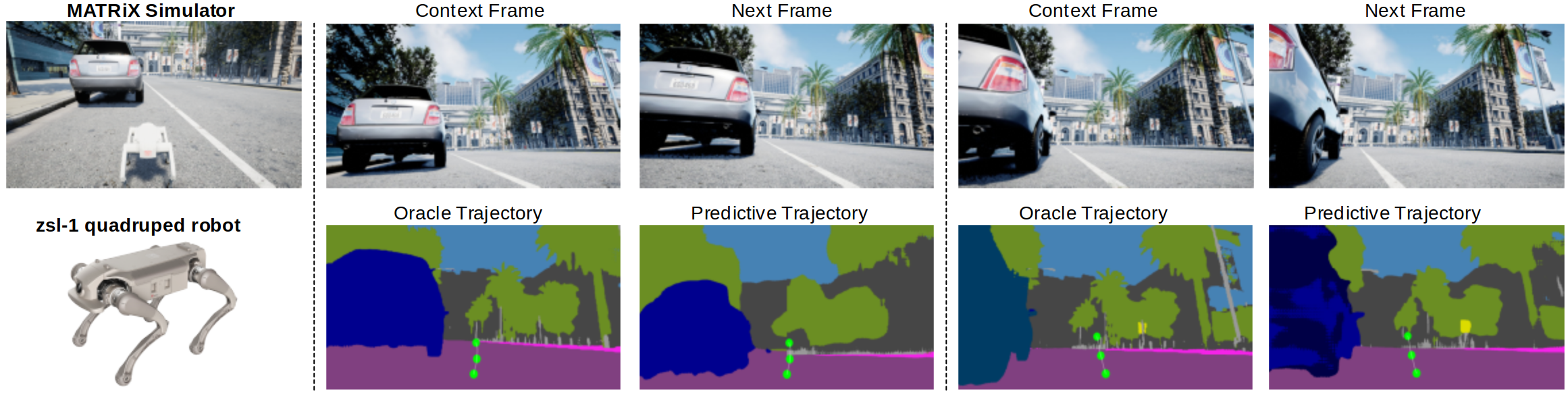}
    \caption{\textbf{Qualitative deployment results in the MATRiX simulator.} The left column illustrates the simulation environment and the zsl-1 quadruped robot. We integrate the forecasted features of DF$^3$ with ViPlanner to perform trajectory prediction. The green points visualize the predicted trajectory waypoints. This integration enables the quadruped robot to accurately anticipate environmental dynamics and plan safe navigation paths.}
    \label{fig:vis_traj}
\end{figure*}

\subsection{Qualitative Results}
Fig.~\ref{fig:vis_seg} compares the semantic segmentation results derived from our forecasted features against the Oracle. DF$^3$ successfully predicts the future states of dynamic scenes despite the absence of a heavy spatial decoder. The model accurately models the motion of dynamic objects and captures their future positions, such as the rapidly moving car in row (a) and the crossing cyclists in row (c). The structural layout of the static background also remains highly consistent with the target frame. This confirms that our MACF mechanism effectively propagates historical temporal dynamics into the latent space.

We acknowledge that the forecasted features exhibit certain limitations regarding fine-grained fidelity. As visualized, object boundaries appear slightly coarse with occasional edge noise, and thin structures frequently lose sharpness. This spatial over-smoothing is an inherent compromise of predicting strictly within the downsampled feature space of a frozen ViT. However, we argue that for autonomous navigation systems, computational efficiency significantly outweighs the need for pixel-perfect visual precision. High-frequency boundary details offer marginal benefits for downstream decision making, provided that the macroscopic motion of dynamic objects is correctly modeled.

To further validate this hypothesis and highlight the zero-shot generalization capabilities of our framework, we deploy the DF$^3$ directly into the MATRiX simulator~\cite{zsibot2026matrix}, an advanced simulator for quadruped robots, to control its built-in zsl-1 robot. Remarkably, the model trained exclusively on automotive perspectives from Cityscapes drives the trajectory planning of a quadruped robot without fine-tuning. We integrate our method with ViPlanner~\cite{viplanner} to compare navigation trajectories generated from our forecasted features against those derived from Oracle future frames. As visualized in Fig.~\ref{fig:vis_traj}, the results naturally appear coarser with slight spatial misalignments. This visual degradation is expected given the substantial domain gap between the Cityscapes automotive training data and the inherently jittery, ground-level locomotion of the quadruped robot. However, this visual disparity actively reinforces our core argument. The strong spatial alignment between our predictive trajectory and the Oracle trajectory proves that DF$^3$ successfully extracts the essential motion dynamics required for safe planning regardless of the viewpoint. This successful cross-embodiment transfer confirms that our compact latent representations provide highly robust and actionable spatial information for real-time robotic control of diverse physical agents.


\section{Conclusions}
In this paper, we proposed Decoder-Free Feature Forecasting (DF$^3$), a highly efficient latent world modeling framework. By eliminating the computational bottleneck of heavy decoders, DF$^3$ directly forecasts future states and derives task outputs. Our approach leverages learnable spatial queries injected into a frozen vision foundation model, coupled with a lightweight, motion-aware context fusion mechanism. This allows the queries to explicitly align historical tokens and forecast future features. Experiments demonstrate that DF$^3$ achieves state-of-the-art forecasting performance while significantly reducing computational overhead, providing a scalable foundation for real-time decision-making in robotic systems.

\noindent\textbf{Limitations.} While DF$^{3}$ achieves highly efficient latent world modeling, it is currently limited to observation-only future forecasting. The present framework does not support action-conditional prediction, which is a crucial component for closed-loop control and planning in interactive environments. In future work, we will explore incorporating action conditions into our motion-aware context fusion mechanism and adopting lighter vision backbones, enabling the model to simulate the consequences of specific robot actions more rapidly directly within the latent space.

\bibliographystyle{IEEEtran}
\bibliography{IEEEabrv,main}

\end{document}